\def\year{2019}\relax
\documentclass[letterpaper]{article} 
\usepackage{aaai19}  
\usepackage{times}  
\usepackage{helvet}  
\usepackage{courier}  
\usepackage{graphicx}  

\usepackage[hidelinks]{hyperref}
\usepackage{booktabs}
\usepackage{tabularx}
\usepackage{amsmath}
\usepackage{csquotes}
\usepackage{pgfplots}

 \usepackage{natbib}
 \renewcommand\cite{\citep}	

\begin{document}
%
\title{Jointly Learning to Label Sentences and Tokens}

\author{AAAI Press\\
Association for the Advancement of Artificial Intelligence\\
2275 East Bayshore Road, Suite 160\\
Palo Alto, California 94303\\
}
\author{Marek Rei \\
 The ALTA Institute \\
 Computer Laboratory \\
 University of Cambridge \\
 United Kingdom\\
 {\tt marek.rei@cl.cam.ac.uk} \\\And
 Anders S{\o}gaard \\
 CoAStaL DIKU\\
 Department of Computer Science\\
 University of Copenhagen\\
 Denmark \\
 {\tt soegaard@di.ku.dk} \\}

\maketitle
\begin{abstract}
Learning to construct text representations in end-to-end systems can be difficult, as natural languages are highly compositional and task-specific annotated datasets are often limited in size.
Methods for directly supervising language composition can allow us to guide the models based on existing knowledge, regularizing them towards more robust and interpretable representations.
In this paper, we investigate how objectives at different granularities can be used to learn better language representations and we propose an architecture for jointly learning to label sentences and tokens.
The predictions at each level are combined together using an attention mechanism, with token-level labels also acting as explicit supervision for composing sentence-level representations.
Our experiments show that by learning to perform these tasks jointly on multiple levels, the model achieves substantial improvements for both sentence classification and sequence labeling.


\end{abstract}

\section{Introduction}

Language composition is a core requirement for many Natural Language Processing tasks, as it allows systems to construct the meaning of a phrase or sentence by combining individual words.
Many languages are highly compositional and the correct predictions often depend on long sequences of context, therefore it is crucial that the learned composition functions are dynamic and accurate.
LSTMs \citep{Hochreiter1997} have become a common approach for constructing text representations, providing a method for iteratively combining word embeddings and learning practical composition functions \citep{Sutskever2014,Huang2015,Kim2016,Melis2017}. 
Convolutional networks have also been explored as an alternative, operating over fixed window sizes and allowing for more computation to be parallelized \citep{Kim2014,Dauphin2017,Gehring2017}.
Each of these can be further extended with an attention-based architecture, giving the model more flexibility and allowing it to dynamically decide which areas of the text should receive more focus for a given task \citep{Bahdanau2015,Yang2016a}.

Most contemporary neural architectures are trained end-to-end, expecting the models to discover the necessary functions for language composition and attention. 
In practice, the variability in natural language is very large and the available annotated datasets are often too small to learn everything automatically.
Furthermore, the patterns discovered in the data might not always correspond to the behaviour that we expect or desire our models to exhibit.
Therefore, we would benefit from having methods for directly supervising the composition functions based on previous knowledge and available annotation, while still taking advantage of the end-to-end supervision signal of the main task.
Recent work has shown that alignment annotation can be used to guide the attention function in neural machine translation models \citep{liu2016neural}.
However, similar approaches could also be extended beyond alignment, making them applicable to learning better composition functions for a wide variety of tasks.

In this paper, we investigate how supervised objectives at different granularities can be combined in order to learn better overall language representations and composition functions.
The proposed model uses a self-attention mechanism for constructing sentence-level representations for text classification, learning to dynamically control the focus at each word position.
We modify the attention component to also behave as a token labeling system, allowing us to directly supervise the attention values based on existing token-level annotation.
Auxiliary objectives using language modeling are also investigated, which regularize the composition functions towards the language style of the training domain.

The joint labeling objective encourages the model to apply more attention to the same areas as the human annotators, making the system more robust to noise in the training data and the model behaviour more intelligible for human users.
The token-level predictions directly reflect the internal decisions of the composition model, making them well-suited for interpreting the model and analysing the output.
The token labeling component itself also benefits from the sentence-level objective, as this performs task-specific regularization and compensates for noisy or missing labels.
The approach is evaluated on three different language analysis tasks where labeling can be performed either on sentences or tokens: grammatical error detection, uncertainty detection and sentiment detection.
The joint labeling architecture is able to return predictions for both full sentences and individual tokens, and the combined model is shown to improve performance on both levels of granularity.

\section{Model Architecture}
\label{sec:model}

The text classification model takes as input a sequence of tokens, such as a sentence, and first predicts a relevance score for each token.
These scores are used as weights in a self-attention mechanism that constructs a combined representation and predicts an overall label for the input text.
The attention module is optimised as a sequence labeling system, explicitly teaching it how to recognize the most important areas in the text.
This ties together the label predictions on different levels, encouraging the objectives to work together and improve performance on both tasks.
The architecture is based on the zero-shot sequence labeling framework by \citet{rei2018zero} which we extend with additional objectives and joint supervision on multiple levels.
We will first describe the core architecture of the model and then provide details on different objective functions for optimization.


\subsection{Language Composition Model}
\label{sec:modelcomp}

The tokens given as input to the system are first broken down into individual characters, and vector representations are constructed for each word by composing character embeddings with a bi-directional LSTM.
The last hidden vectors from each of the LSTMs are then concatenated and passed through a layer with \textit{tanh} activation.
Next, we combine the character-based representation $m_i$ with a pre-trained word embedding $w_i$ by concatenating them together, following \citet{Lample2016}.
This allows the model to learn character-level features while also taking advantage of high-quality embeddings trained on large unannotated corpora.
Vector $x_i$ will now represent the $i$-th word in downstream modules of sentence composition and analysis. 
During training, all of these components are updated end-to-end, including word embeddings, character embeddings and the character-level LSTMs.

The word representations $x_i$ are given as input to a bi-directional LSTM which operates over the words in the sentence, moving in both directions:
\begin{align}
\overrightarrow{h_i} =& LSTM(x_i, \overrightarrow{h_{i-1}})\\
\overleftarrow{h_i} =& LSTM(x_i, \overleftarrow{h_{i+1}})
\end{align}

\noindent The hidden states from each direction are concatenated at every token position and passed through a linear layer, resulting in vector $h_i$ that represents the word at position $i$, but is also conditioned on the whole surrounding sentence:
\begin{align}
\widetilde{h_i} =& [\overrightarrow{h_i};\overleftarrow{h_i}]\\
h_i =& tanh(W_h \widetilde{h_i} + b_h)
\label{eq:h}
\end{align}

\begin{figure}[t]
	\centering
	\includegraphics[width=0.6\linewidth]{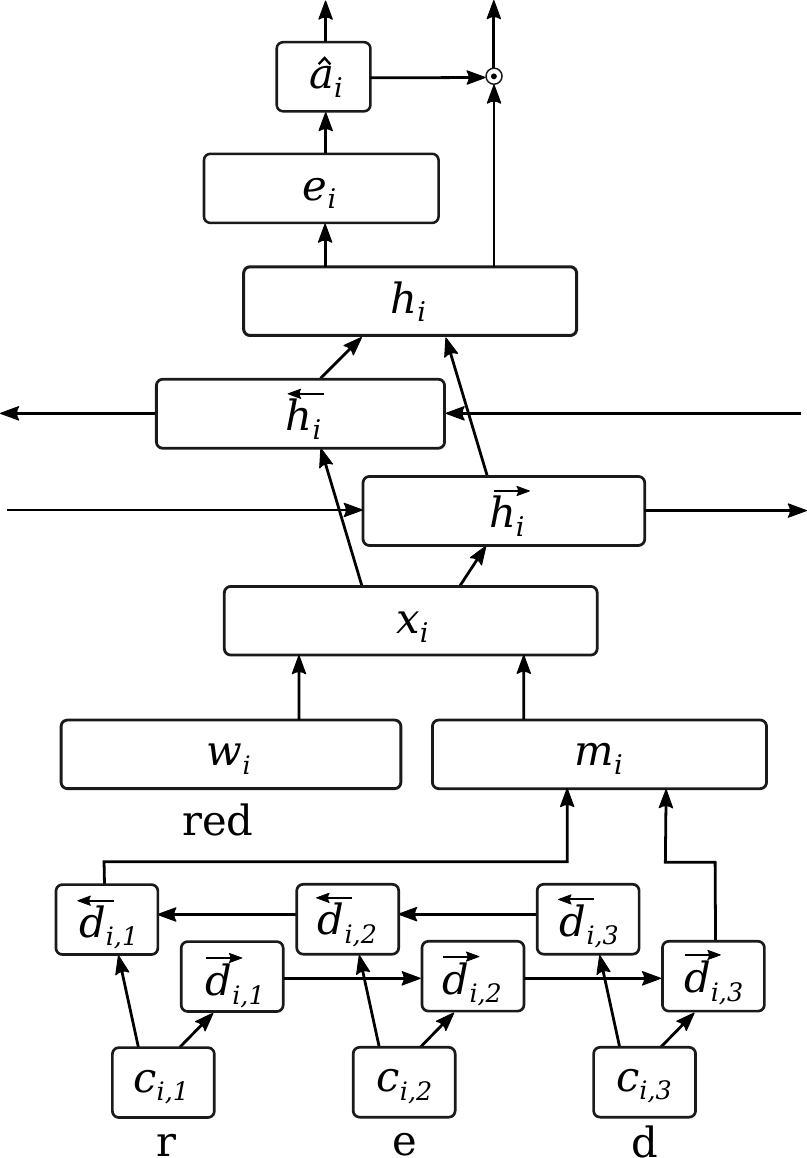}
	\caption{The model architecture for one specific token position, taking the word \textit{'red'} as input and returning a token-level prediction $\hat{a}_i$ together with a weighted vector representation.}
	\label{fig:network}
\end{figure}

Labels are then predicted for each word by passing $h_i$ through a non-linear layer and then to a single output neuron with sigmoid activation:
\begin{align}
e_i =& tanh(W_e h_i + b_e)\\
\hat{a}_i =& \sigma(W_a e_i + b_a)
\end{align}

The value of $\hat{a}_i$ is between $0$ and $1$, representing the confidence of the $i$-th token having a positive label.
In order to predict a label for the overall sentence, we construct a sentence representation using a self-attention mechanism, similar to \citet{Yang2016a}, with the token-level predictions functioning as attention weights:
\begin{equation}
\widetilde{a}_i = \frac{\hat{a}_i}{\sum_{k=1}^N \hat{a}_k}
\end{equation}

\noindent where $\widetilde{a}_i$ is the attention weight, normalized to sum up to $1$ over all values in the sentence. 
These weights are then used for combining the context-conditioned hidden representations from Equation \ref{eq:h} into a single sentence representation $s$:
\begin{equation}
s = \sum_{i=1}^{N} \widetilde{a}_i h_i
\end{equation}

Finally, we return a sentence-level prediction based on this representation:
\begin{equation}
y = \sigma(W_y tanh(W_{\widetilde{y}} s + b_{\widetilde{y}}) + b_y)
\end{equation}

The architecture takes advantage of soft-attention \cite{Shen2016}, where the weights are calculated using the logistic function instead of an exponential function. This provides a smoother distribution over the sentence and allows the same value to also represent the token-level prediction.
The aim of an attention mechanism is to allow the model to choose which areas of the sentence it should focus on. By using token-level predictions as attention weights, we are explicitly tying together the predictions on both levels. For example, in order to decide whether the sentence has positive sentiment, the model will assign higher attention weights to tokens that it has identified as being positive.

The baseline model is only optimized as a sentence-level classifier, minimizing the mean squared error between the predicted sentence-level score $\hat{y}^{(t)}$ and the gold-standard sentence label $y^{(t)}$:
\begin{equation}
L_{sent} = \sum_t (\hat{y}^{(t)} - y^{(t)})^2
\end{equation}

The experiments in this paper focus on binary classification, but the model could also be extended for multi-label classification tasks in the future.
In the next sections, we will explore different auxiliary objective functions that improve the optimization of this architecture.

\subsection{Supervised Attention}

The model uses shared values for token-level predictions and attention weights.
When optimizing for sentence classification, the network will learn to focus on informative areas of the sentence by itself.
However, for some tasks we can take advantage of available token-level annotation and present this as an additional training signal.
By providing token-level supervision to the sentence classification model, we are able to teach it to focus on the most relevant areas and thereby improve the quality of sentence representations along with overall performance.

To achieve this, we add an extra objective function for directly optimizing the predicted word score $\hat{a}_i$:
\begin{equation}
L_{tok} = \sum_t \sum_i (\hat{a}^{(t)}_i - a^{(t)}_i)^2
\end{equation}
\noindent where $\hat{a}^{(t)}_i$ is the predicted score for word $i$ in sentence $t$, and $a^{(t)}_i$ is the binary gold-standard label for the same word.
This objective is commonly used in a sequence labeling architecture and it optimizes the model to assign correct labels to every token.
However, since the token-level predictions are directly tied to attention weights in our composition function, this objective also teaches the model to focus on the most relevant areas in the text when composing sentence representations.

\subsection{Language Modeling Objective}

\citet{Rei2017} proposed an auxiliary objective for sequence labeling, where the architecture is also trained as a language model.
This method regularizes the network, while also providing a richer training signal for the language composition functions, specializing them for the given domain and writing style.
The optimization is performed concurrently with the other tasks, using the same training set, which means we are able to improve performance without any additional data.

This objective has not been previously investigated beyond token labeling, therefore we integrate the language modeling objective into our network and investigate whether it also improves performance when composing sentence-level representations. 
Each of the hidden LSTM states is passed through a specialized layer and then mapped to a probability distribution over the vocabulary of words using the softmax function:
\begin{align}
\overrightarrow{q_i} =& tanh(\overrightarrow{W}_q \overrightarrow{h_i} + \overrightarrow{b}_q)\\
\overleftarrow{q_i} =& tanh(\overleftarrow{W}_q \overleftarrow{h_i} + \overleftarrow{b}_q)\\
P(w_{i+1}|\overrightarrow{q_i}) =& softmax(\overrightarrow{W}_{\widetilde{q}} \overrightarrow{q_i} + \overrightarrow{b}_{\widetilde{q}})\\
P(w_{i-1}|\overleftarrow{q_i}) =& softmax(\overleftarrow{W}_{\widetilde{q}} \overleftarrow{q_i} + \overleftarrow{b}_{\widetilde{q}})
\end{align}

\noindent where $\overrightarrow{q_i}$ and $\overleftarrow{q_i}$ are hidden layers that allow the representation to specialize for the language modeling task.
The model is then optimized to predict the next word in the sequence based on the forward-moving LSTM, and the previous word in the sequence based on the backward-moving LSTM, while the combination of both is used for assigning labels to each token:
\begin{equation}
\begin{split}
L_{LM} =& - \sum_{i=1}^{N-1} log(P(w_{i+1}|\overrightarrow{q_i}))\\
           & - \sum_{i=2}^{N} log(P(w_{i-1}|\overleftarrow{q_i}))
\end{split}
\end{equation}

\subsection{Character-level LM Objective}

We further extend the idea of using an auxiliary language modeling objective, and apply it to the character-based representations.
Learning character-level composition functions and embeddings can be difficult, and learning to predict surrounding words provides the model with additional morphological information.
However, the character-LSTMs compose only single words, and using them to predict the next or previous word would only be equivalent to a unigram language model.
In order to provide some additional context to the model, we instead optimize the network to predict words in the sentence based on the character-level hidden states of the surrounding words from both sides.
\begin{align}
\widetilde{g} = [\overrightarrow{d}_{i-1,R};&\overleftarrow{d}_{i-1,1};\overrightarrow{d}_{i+1,R};\overleftarrow{d}_{i+1,1}]\\
g_i &= tanh(\overleftarrow{W}_g \widetilde{g} + b_g)\\
P(w_{i}|g_{i}) &= softmax(\overrightarrow{W}_{\widetilde{g}} g_i + b_{\widetilde{g}})
\end{align}

\noindent where $\overrightarrow{d}_{i,j}$ is the left-to-right character-based representation for character $j$ in word $i$, $\widetilde{g}$ contains the concatenation of the four hidden states from the bi-directional character-level LSTMs from the previous and next word in the sequence. The model is then optimized by minimizing the negative log-likelihood of predicting the middle word in the sequence:
\begin{equation}
L_{char} =- \sum_{i=2}^{N-1} log(P(w_{i}|g_i))\\
\end{equation}

Both the word-level and character-level LM objectives introduce additional parameters to the model, in order to predict the probability distribution over surrounding words. However, these parameters are only required during training and can be omitted during testing, resulting in the same model architecture as the baseline, with the same number of parameters.

\subsection{Attention Range Objective}

Finally, we also consider a secondary method for joining the sentence-level objective with token labeling. 
For this, we make the assumptions that 1) only some, but not all, tokens in the sentence can have a positive label, and 2) that there are positive tokens in a sentence only if the overall sentence is positive.
We can then construct a loss function that encourages the model to optimize for these constraints:
\begin{equation}
 L_{attn} = \sum_t (min(\hat{a}_i^{(t)}) - 0)^2 + 
             \sum_t (max(\hat{a}_i^{(t)}) - y^{(t)})^2
\end{equation}

\noindent where $min(\hat{a}_i^{(t)})$ is the minimum value of all the attention weights in sentence $t$ and $max(\hat{a}_i^{(t)})$ is the corresponding maximum value.
The first part of the equation pushes the minimum unnormalized attention weights in a sentence towards 0, satisfying the constraint that all tokens in a sentence should not have a positive token-level label.
The second component then optimizes for the maximum unnormalized attention weight in a sentence to be equal to the gold label for that sentence, which is either $0$ or $1$, incentivizing the network to only assign large attention weights to tokens in positive sentences.
While these assumptions are not always true, they provide a method of connecting the sentence- and token-level optimization even when the token-level annotation is noisy or missing.

\begin{table*}[t]
\centering
\setlength\tabcolsep{9pt}
\begin{tabular}{l|ccccc|ccccc} \toprule
 & \multicolumn{5}{c|}{{\small CoNLL 2010}} & \multicolumn{5}{c}{{\small FCE}} \\
  & {\small DEV $F_1$} & {\small Acc} & {\small P} & {\small R} & {\small $F_1$} & {\small DEV $F_1$} & {\small Acc} & {\small P} & {\small R} & {\small $F_1$} \\ \midrule
{\small BiLSTM-LAST} & 90.17 & 94.95 & 85.66 & 81.87 & 83.67 & 84.69 & 77.75 & 78.55 & 92.55 & 84.95 \\ 
{\small BiLSTM-ATTN} & 89.88 & 94.98 & 85.34 & 82.68 & 83.87 & 84.96 & 78.12 & 78.75 & \textbf{92.87} & 85.21 \\
{\small BiLSTM-JOINT} & \textbf{91.30} & \textbf{95.97} & \textbf{87.63} & \textbf{86.76} & \textbf{87.17} & \textbf{86.14} & \textbf{80.08} & \textbf{82.27} & 90.14 & \textbf{86.01} \\ \bottomrule
\end{tabular}
\caption{
Sentence classification results on CoNLL 2010 and FCE datasets. {\small BiLSTM-LAST} uses the last hidden states; {\small LSTM-ATTN} uses the attention-based composition while only optimizing for sentence classification; {\small BiLSTM-JOINT} is the full multi-level model, receiving supervision on both sentences and tokens.
}
\label{tab:results1}
\end{table*}

\section{Datasets}


We evaluate the joint labeling framework on three different tasks and datasets.
The \textbf{CoNLL 2010} shared task \cite{Farkas2010} dataset investigates the detection of uncertain language, also known as hedging. Speculative language is a common tool in scientific writing, allowing scientists to guide research beyond the evidence, and roughly 19.44\% of sentences in  biomedical papers contain hedge cues \cite{Vincze2008}.
Automatic detection of these cues is important for downstream tasks such as information extraction and literature curation, as typically only definite information should be extracted and curated.

The shared task consisted of two separate subtasks: 1) detection of uncertainty in text by predicting a binary label for a sentence, and 2) detection of the location of individual cue tokens and their semantic scope.
The dataset is annotated for both hedge cues (keywords indicating uncertainty) and scopes (the area of the sentence where the uncertainty applies). In our experiments, the models aim to detect hedge cues on the token level and the presence of speculative language on the sentence level. The joint labeling framework encourages the model to classify sentences based on detected hedge cues, while also using the sentence-level objective to improve the cue detection itself.

We also evaluate the model error detection, where the goal is to identify tokens which need to be edited in order to produce a grammatically correct sentence. 
This task is an important component in automated systems for language teaching and assessment, with recent work focusing on error detection as a supervised sequence labeling task \cite{Rei2016,Kaneko2017,Rei2017}.
In turn, classifying sentences as correct or ungrammatical is necessary for developing tutoring systems for language learners \cite{Andersen2013,Daudaravicius2016} and detecting errors in machine-generated text.

For error detection on both levels, we use the First Certificate in English (\textbf{FCE}, \citet{Yannakoudakis2011}) dataset, containing error-annotated short essays written by language learners. 
\citet{Rei2016} converted the original error correction annotation to a sequence labeling dataset, which we use in our experiments.
The model is optimized to detect incorrect sentences, while also identifying the location of the error tokens, with the joint model combining both objectives together.

Finally, we convert the Stanford Sentiment Treebank (\textbf{SST}, \citet{Socher2013}) to a sequence labeling format, in order to evaluate on the task of sentiment detection. Sentiment analysis is generally a three-way classification task, whereas the attention framework is designed for binary classification. Therefore, we construct two separate binary tasks from this dataset -- the detection of positive and negative sentiment.

In order to assign sentiment labels to individual tokens, we traverse the treebank annotation bottom-up. 
The sentiment label is assigned to the minimum span of tokens containing that sentiment, but can be overwritten by subsuming phrases of up to length 3 with the opposing sentiment. For example, in the phrase \textit{'This movie is good'}, only the token \textit{'good'} will be labeled as positive, whereas in the phrase \textit{'This movie is not good'} the tokens \textit{'not good'} are labeled as negative.
The model is optimized to detect the presence of positive/negative sentiment on the sentence level, while also labeling the individual tokens with the corresponding sentiment.

\begin{table*}[t]
\centering
\setlength\tabcolsep{9pt}
\begin{tabular}{l|ccccc|ccccc} \toprule
 & \multicolumn{5}{c|}{{\small SST-neg}} & \multicolumn{5}{c}{{\small SST-pos}} \\
 & {\small DEV $F_1$} & {\small Acc} & {\small P} & {\small R} & {\small $F_1$} & {\small DEV $F_1$} & {\small Acc} & {\small P} & {\small R} & {\small $F_1$} \\ \midrule
{\small BiLSTM-LAST} & 88.27 & 84.19 & 87.59 & 90.37 & 88.95 & 93.82 & 88.76 & 90.72 & \textbf{96.85} & 93.67 \\ 
{\small BiLSTM-ATTN} & 88.99 & 85.32 & 88.18 & \textbf{91.43} & 89.77 & 93.95 & 89.33 & 91.47 & 96.59 & 93.96 \\
{\small BiLSTM-JOINT} & \textbf{91.62} & \textbf{88.42} & \textbf{92.45} & 90.98 & \textbf{91.71} & \textbf{96.37} & \textbf{93.30} & \textbf{96.22} & 95.97 & \textbf{96.09} \\ \bottomrule
\end{tabular}
\caption{Sentence classification results on Stanford sentiment treebank, separated into negative and positive sentiment detection.}
\label{tab:results2}
\end{table*}

\section{Implementation Details}

We combine all the different objective functions together using weighting parameters. 
This allows us to control the importance of each objective and prevent the model from over-specializing to auxiliary goals.
The final objective that we minimize during training is then:
\begin{equation}
\begin{split}
L =& \Lambda_{sent} \cdot L_{sent} + \Lambda_{tok} \cdot  L_{tok}\\
           & + \Lambda_{LM}  \cdot L_{LM} + \Lambda_{char} \cdot  L_{char} \\
           & + \Lambda_{attn} \cdot  L_{attn}
\end{split}
\end{equation}

For the baseline system we set $\Lambda_{sent}=1$ and all the other weights to $0$, optimizing only for the sentence-level label prediction.
When using the full system, we use $\Lambda_{sent}=\Lambda_{tok}=1$, $\Lambda_{LM}=\Lambda_{char}=0.1$ and $\Lambda_{attn}=0.01$. The sentence- and token-labeling objectives are part of the main task and therefore set to $1$; the language modeling objective weights were chosen following \citet{Rei2017}, and $\Lambda_{attn}$ was chosen based on experiments on the development set.

Tokens were lowercased, while the character-level component receives input with the original capitalization.
Word embeddings were set to size 300, pre-loaded from publicly available Glove embeddings \cite{Pennington} and fine-tuned during training.
The word-level LSTMs are size 300 and character-level LSTMs size 100; the hidden combined representation $h_i$ was set to size 200; the attention weight layer $e_i$ was set to size 100.

The model was optimized using AdaDelta \cite{Zeiler2012} with learning rate $1.0$.
The network weights were randomly initialized using the uniform Glorot initialization method \cite{Glorot2010}.
Dropout \cite{Srivastava2014a} with probability $0.5$ was applied to word representations $w_i$ and the composed representations $h_i$ after the LSTMs.
Training was stopped if performance on the development set had not improved for 7 epochs and the best model was used for evaluation.
The code for running these experiments will be made publicly available.\footnote{http://www.marekrei.com/projects/mltagger}

\section{Sentence Classification Experiments}
\label{sec:sentence}

We first evaluate the architectures on different sentence classification tasks. Table \ref{tab:results1} contains results for detecting speculative language and grammatical errors on the sentence level. Table \ref{tab:results2} presents the results for the two subtasks of sentiment classification.
Table \ref{tab:results3} contains fine-grained results, evaluating each of the proposed objectives in isolation. This acts as an ablation test, allowing us to determine whether the individual modifications benefit overall performance. 

{\small BiLSTM-LAST} is our baseline architecture, commonly used for similar text composition tasks \cite{tang2015document,neelakantan2016neural}. 
It processes the input with a bi-directional LSTM, concatenates the hidden states  from both directions and uses these to predict the sentence-level label.
{\small BiLSTM-ATTN} is the architecture using attention, based on \citet{Yang2016a}, which is optimized only using the sentence-level objective.
{\small BiLSTM-JOINT} is the full multi-level model, composing the sentences with self-attention and incorporating all of the proposed objectives during the training process.

Analyzing the results, we find that the attention-based architecture itself already gives consistent improvements, with {\small BiLSTM-ATTN} outperforming {\small BiLSTM-LAST} in all settings. The self-attention framework allows the model to dynamically choose the areas of the sentence where to focus, delivering better results on all datasets.
The proposed auxiliary objectives also provide consistent improvements in $F_1$ score when combined with the {\small BiLSTM-ATTN} model for sentence classification.
Integrating the token-labeling objective with the attention gives the biggest improvements overall, 
showing that explicitly supervising the attention function allows the model to learn better sentence representations. With over $25\%$ error reduction in $F_1$, the token-level objective is particularly beneficial for the sentiment analysis datasets, as it teaches the model to detect key phrases in the text.

\begin{table}[t]
\centering
\setlength\tabcolsep{7pt}
\begin{tabular}{l|cccc} \toprule
 & {\small C'10} & {\small FCE} & {\small SST-neg} & {\small SST-pos} \\ \midrule
{\small BiLSTM-LAST} & 83.67 & 84.95 & 88.95 & 93.67 \\ \midrule
{\small BiLSTM-ATTN} & 83.87 & 85.21 & 89.77 & 93.96 \\
 +token & 85.90 & 85.41 & \textbf{91.97} & 95.82 \\
 +{\small LM} word & 85.70 & 85.48 & 89.71 & 94.02 \\
 +{\small LM} char & 85.08 & 85.35 & 89.67 & 93.79 \\
 +attn cost & 83.95 & 85.21 & 89.81 & 93.79 \\ \midrule
{\small BiLSTM-JOINT} & \textbf{87.17} & \textbf{86.01} & 91.71 & \textbf{96.09} \\ \bottomrule
\end{tabular}
\caption{Comparing sentence classification performance when each of the auxiliary objective functions is added to {\small BiLSTM-ATTN} in isolation.}
\label{tab:results3}
\end{table}

The findings also indicate that the language modeling objective is indeed beneficial in the text classification setting. By training the BiLSTM components as separate language models, we are providing the model with a richer optimization signal and a natural method of regularization which automatically matches the given corpus style. 
Applying a language modeling objective on the characters also delivers a separate improvement in performance. Character-based neural models need sufficient amounts of data to train, whereas human-annotated training resources can be very limited in size. By including this auxiliary objective, we are able to optimize the model to learn informative character-level features without requiring additional training data.
The improvements of the attention range objective varied depending on the dataset, but we found it to give small yet consistent improvements when added on top of other objectives in the combination system. Incentivizing the model to make token-level predictions in a suitable range is also beneficial in settings when only sentence-level annotation is available.

Finally, combining all the objectives together gave the best and most consistent overall performance across the board. The {\small BiLSTM-JOINT} model achieves the highest $F_1$ score on three out of four datasets. The exception is negative sentiment detection, only because the token-level objective proved to be particularly important on that dataset.
By teaching the model to focus in the right areas in the text and predict unseen words in the context, we are able to obtain a more robust and accurate sentence classification system.

\section{Token Labeling Experiments}

In this section, we investigate how the {\small BiLSTM-JOINT} architecture performs as a token-level sequence labeler. We use the same model and the same training objectives, except for a small difference in the early stopping criterion: in previous experiments, we stopped training based on the sentence-level performance on the development data, whereas now we use the corresponding token-level measure.  In general, we found that the model requires more epochs before reaching its optimal performance on the sequence labeling tasks, likely due to the datasets having relatively fewer unique instances for sentence-level classification.

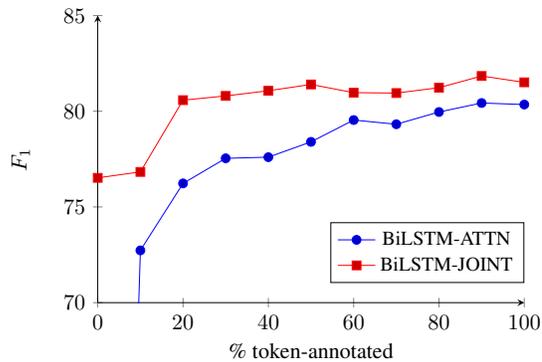
\begin{figure}[t]
		\centering
		\begin{tikzpicture}[scale=0.83]
		\begin{axis}[axis x line=left, axis y line=left, xlabel=\% token-annotated, ylabel= $F_1$, legend style={at={(1,0.28)},
			anchor=north east, legend columns=1}, xmin=0, xmax=100, ymin=70, ymax=85, width=\columnwidth, height=0.85*\axisdefaultheight] 
		\addplot+[] coordinates
		{(0,1.33) (10,72.73) (20,76.23) (30,77.54) (40,77.60) (50,78.40) (60,79.54) (70,79.32) (80,79.96) (90,80.43) (100,80.35)}
		node[] {};
		\addlegendentry{\small BiLSTM-ATTN}
		
		\addplot+[] coordinates
		{(0,76.52) (10,76.83) (20,80.58) (30,80.80) (40,81.07) (50,81.40) (60,80.97) (70,80.95) (80,81.23) (90,81.84) (100,81.51)}
		node[] {};
		\addlegendentry{\small BiLSTM-JOINT}
		
		\end{axis}
		\end{tikzpicture}
	\caption{Sequence labeling $F_1$ on CoNLL-10 cue detection when varying the amount of training data that is token-annotated.}
	\label{fig:gradient1}
\end{figure}

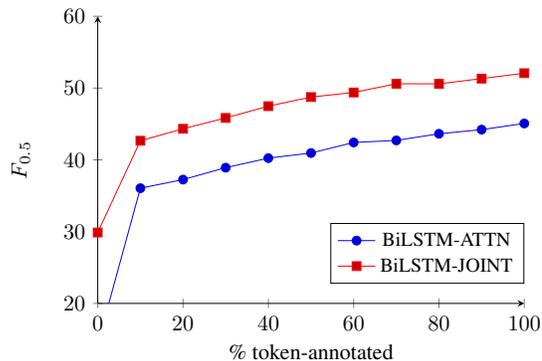
\begin{figure}[t]
		\centering
		\begin{tikzpicture}[scale=0.83]
		\begin{axis}[axis x line=left, axis y line=left, xlabel=\% token-annotated, ylabel= $F_{0.5}$, legend style={at={(1,0.28)},
			anchor=north east, legend columns=1}, xmin=0, xmax=100, ymin=20, ymax=60, width=\columnwidth, height=0.85*\axisdefaultheight]  
        %
		\addplot+[] coordinates
		{(0,14.37) (10,36.05) (20,37.25) (30,38.91) (40,40.24) (50,40.95) (60,42.42) (70,42.71) (80,43.63) (90,44.21) (100,45.07)}
		node[] {};
		\addlegendentry{\small BiLSTM-ATTN}
		
		\addplot+[] coordinates
		{(0,29.88) (10,42.66) (20,44.33) (30,45.84) (40,47.48) (50,48.75) (60,49.39) (70,50.60) (80,50.59) (90,51.32) (100,52.07)}
		node[] {};
		\addlegendentry{\small BiLSTM-JOINT}
		
		\end{axis}
		\end{tikzpicture}
	\caption{Sequence labeling $F_{0.5}$ on FCE error detection when varying the amount of token-annotated training data.}
	\label{fig:gradient2}
\end{figure}

While regular sequence labeling models are trained using token-annotated examples, our model is able to also take advantage of sentence-annotated data.
Collecting human annotation on the sentence level can be considerably easier and cheaper for many tasks, compared to labeling individual tokens.
Experiments in Figures \ref{fig:gradient1} and \ref{fig:gradient2} measure sequence labeling performance as the percentage of available token-annotated sentences is varied, with the remaining examples only having sentence-level labels. 
In the absence of token-level annotation, our model will still continue to optimize the sequence labeling components -- the sentence classification objective encourages the model to focus on relevant tokens in the sentence, and the attention range objective adjusts the output values into the correct scale.
Remarkably, this training signal is strong enough to achieve reasonable sequence labeling performance even without any token-annotated data (0\% on the scale), learning to label individual words based only on the sentence-level annotation.
Using only 20\% of the token-level annotation, the joint model achieves performance comparable to a regular sequence tagging model trained on the full dataset. The system also benefits from the auxiliary objectives when all the token-level annotation is available, performing $1.3\%$ better on speculative language detection and a substantial $7\%$ better on error detection.

\begin{table*}[t]
\centering
\setlength\tabcolsep{6.7pt}
\begin{tabular}{l|ccc|ccc|ccc|ccc} \toprule
 & \multicolumn{3}{c|}{{\small FCE}} & \multicolumn{3}{c|}{{\small CoNLL14-TEST1}} & \multicolumn{3}{c|}{{\small CoNLL14-TEST2}} & \multicolumn{3}{c}{{\small JFLEG}} \\
& {\small P} & {\small R} & {\small $F_{0.5}$} & {\small P} & {\small R} & {\small $F_{0.5}$} & {\small P} & {\small R} & {\small $F_{0.5}$} & {\small P} & {\small R} & {\small $F_{0.5}$} \\ \midrule
Rei (2017) & 58.88 & \textbf{28.92} & 48.48 & 17.68 & \textbf{19.07} & 17.86 & 27.62 & \textbf{21.18} & 25.88 & - & - & -\\
{\small BiLSTM-ATTN} & 60.73 & 22.33 & 45.07 & 21.69 & 11.42 & 18.16 & 34.13 & 12.76 & 25.22 & 69.86 & 19.32 & 45.74\\
{\small BiLSTM-JOINT} & \textbf{65.53} & 28.61 & \textbf{52.07} & \textbf{25.14} & 15.22 & \textbf{22.14} & \textbf{37.72} & 16.19 & \textbf{29.65} & \textbf{72.53} & 25.04 & \textbf{52.52}\\ \bottomrule
\end{tabular}
\caption{Sequence labeling results on error detection datasets.
Comparing the best system from \citet{Rei2017}, {\small BiLSTM-ATTN} supervised only for sequence labeling, and {\small BiLSTM-JOINT} optimized with all the auxiliary objectives.
}
\label{tab:results4}
\end{table*}

Table \ref{tab:results4} investigates error detection performance in more detail, using the {\small BiLSTM-JOINT} model trained on FCE and evaluating it on external error detection datasets.
The CoNLL 2014 shared task dataset \cite{Ng2013a} contains 1,312 sentences, written by higher-proficiency learners on more technical topics. They have been manually corrected by two separate annotators, and we report results on each of these annotations.
JFLEG \cite{Napoles2017} contains a broad range of language
proficiency levels and focuses more on fluency edits, making the text more native-sounding, in addition to grammatical corrections.
We use $F_{0.5}$ as the main evaluation measure for error detection -- high precision is more important for practical error detection applications, therefore $F_{0.5}$ was established as the main measure by the CoNLL 2014 shared task.

We compare our system to the sequence labeling model by \citet{Rei2017}, which currently has the best reported error detection results on FCE and CoNLL14 when using the dedicated training set.\footnote{Higher results have been reported, but only using various additional annotated datasets.}
The results show substantial improvements and {\small BiLSTM-JOINT} achieves new state-of-the-art results without using any additional training data. Optimizing for sentence composition and language modeling, along with the regular token labeling, provides a more robust system and considerably higher $F_{0.5}$ scores on all the benchmarks. The main impact comes from large improvements in precision, with the additional objectives encouraging the model to learn more informative features, which is ideal for the task of error detection, where high-precision predictions are required.

\section{Related Work}

In recent years, researchers have explored hierarchical neural models for tasks such as part-of-speech tagging \cite{Plank:ea:16}, modeling text coherence \cite{Li:ea:15}, and character-based language modeling \cite{Hermans:Schrauwen:13}. However, none of these consider supervision at both levels of their hierarchical architectures. \citet{Frank:ea:13} present a joint hierarchical model for morphology and syntax, but for unsupervised induction from child-directed speech. 
\citet{zhang2016rationale} make use of sentence labels for document classification, however, predictions from the sentence-level model are given as input to the document model instead of training and composing them jointly. 
Similarly, \citet{ammar2016many} use predictions from a supervised POS tagging component as features for a parsing model.
Recently, \citet{subramanian2018hierarchical} also described a hierarchical model for document-level manifesto analysis based on probabilistic soft logic.

\citet{McDonald:ea:07}~present a related hierarchical model for fine- and coarse-grained sentiment analysis, trained using MIRA, predicting sentiment at both sentence and document levels. They show that the joint model outperforms cascaded models by a wide margin. \citet{Zaidan:Eisner:08} also describe a generative model for fine- and coarse-grained sentiment analysis. 
\citet{Harel:Mannor:11} present an algorithm for learning from what they call multiple {\em outlooks} that is also similar in spirit to our work. Their algorithm takes advantage of the multiple outlooks by matching moments of the empirical distributions to find an optimal mapping between them. However, they do not consider the outlooks at different hierarchical levels.
Most recently, \citet{barrett2018sequence} extended the model described in this paper and used human attention from gaze recordings to train the composition function.


\section{Conclusion}

In this paper, we investigated how supervised objectives at different granularities can be combined in order to learn better overall language representations and composition functions.
The model learns to jointly label text at different granularities, allowing these objectives to benefit from each other. 
We proposed an attention-based model for sentence classification that also behaves as a token labeling system, allowing us to directly supervise the attention values based on existing token-level annotations.
The joint labeling objective encourages the model to apply more attention to the same areas as the human annotators, making the system more robust to noise in the training data and the model behavior more intelligible for human users.
In return, the sentence-level objective provides task-specific regularization for the token labeling component and compensates for noisy or missing labels. 

We also experimented with auxiliary objectives that further assist the model in learning better composition functions that are shared between both tasks.
Training the network to predict surrounding words regularizes the model, while also specializing the language composition functions towards the given domain and writing style.
The language modeling objective can be further extended to character-based representations, providing the character composition model with an additional informative training signal.
Finally, an attention range constraint can be used to connect the labeling objectives on both levels and encourage the attention weights to be in a reasonable range.

The experiments were performed on three different tasks where labeling is needed both on sentences and tokens -- uncertainty detection, sentiment detection and grammatical error detection.
Evaluation of the joint labeling model  showed consistent improvements at labeling both whole sentences and individual tokens, compared to optimizing for these tasks individually. For sequence labeling, the model was also able to use much less training data for comparable results, and even performed reasonably well without any token-level annotations. The joint labeling model with the auxiliary objectives achieved the best and most consistent results on all datasets.

\bibliography{interpsent2}
\bibliographystyle{aaai}

\end{document}